\documentclass[]{spie}  

\usepackage{amsmath,amsfonts,amssymb}
\usepackage[colorlinks=true, allcolors=blue]{hyperref}
\usepackage{booktabs}
\usepackage{multirow}
\usepackage{graphicx}
\usepackage{pifont}

\title{Which Pretext Task Transfers? Self-Supervised Pretraining Objectives for Lung Ultrasound}

\author[a]{Moein Heidari}
\author[a]{Junbo Rao}
\author[a]{Jai Choraria}
\author[b]{Wenjin Chen}
\author[b]{David J. Foran}
\author[a]{Ilker Hacihaliloglu}
\affil[a]{University of British Columbia, Vancouver, BC, Canada}
\affil[b]{Rutgers Cancer Institute of New Jersey, New Brunswick, NJ, USA}

\authorinfo{Send correspondence to M.H.: E-mail: \texttt{moein.heidari@ubc.ca}}

\begin{document} 
\maketitle

\begin{abstract}
Self-supervised learning (SSL) can reduce the need for labelled medical images, but the choice of pretext objective remains unclear for lung ultrasound (LUS). Contrastive learning, masked reconstruction, and joint-embedding predictive architectures (JEPA) differ in the space in which their targets are defined, yet existing ultrasound studies compare them under different corpora, backbones, and evaluation protocols. We compare these three objective families using the same encoder backbone, pretraining corpus, optimisation schedule, and frozen-evaluation protocol. Encoders are pretrained on COVID-BLUeS LUS videos and evaluated with linear, $k$NN, and attentive probes at 5\%, 10\%, 50\%, and 100\% label budgets. Evaluation is performed on POCUS using patient-level five-fold cross-validation and on the independently acquired Mendeley-Uganda dataset, which is excluded from both pretraining and probe fitting. At the full label budget under linear probing, VideoMAE and V-JEPA achieve $66.5 \pm 13.1$ and $65.4 \pm 11.7$ balanced accuracy on POCUS, while MoCo achieves $42.1 \pm 1.2$. On Mendeley-Uganda, the ranking reverses: MoCo performs best at $62.7 \pm 1.0$, followed by VideoMAE at $53.8 \pm 2.8$, while V-JEPA falls near chance at $35.1 \pm 4.9$. These results show that POCUS probe accuracy alone does not identify the objective that transfers best across datasets. We also outline planned representation-level analyses to examine this reversal. Code is publicly available at \href{https://github.com/moeinheidari7829/LUSVideoSSL}{https://github.com/moeinheidari7829/LUSVideoSSL}.
\end{abstract}


\section{INTRODUCTION}
\label{sec:intro}  

Lung ultrasound (LUS) has become a first-line modality for bedside assessment of pulmonary pathology, offering real-time acquisition without ionising radiation. Diagnosis is driven by a compact set of sonographic patterns: A-lines indicate normally aerated parenchyma, B-lines indicate interstitial syndrome, and consolidation and pleural effusion manifest as changes in tissue texture.\cite{born2021accelerating,roy2020deep} A subset of these patterns is inherently dynamic, most notably lung sliding, whose absence is diagnostic of pneumothorax and which is observable only across frames. Interpretation consequently depends on both fine spatial detail and temporal behaviour, and remains strongly operator-dependent. Expert annotation is costly, and publicly available labelled LUS collections are orders of magnitude smaller than the corpora used to train contemporary vision models.

Self-supervised learning (SSL) addresses this constraint by learning representations from unlabelled data, and three families dominate current practice. Contrastive methods enforce invariance between augmented views of the same sample under an InfoNCE objective.\cite{chen2020simple,chen2021empirical} Masked autoencoders reconstruct withheld pixels from a sparse visible context.\cite{he2022masked,tong2022videomae} Joint-embedding predictive architectures (JEPA) instead regress the \emph{representation} of a withheld region from visible context, so that the prediction target is a learned feature rather than an observed pixel.\cite{assran2023self,assran2025v} These objectives are not interchangeable: they impose distinct inductive biases on what the encoder is permitted to discard. Pixel-space reconstruction allocates capacity to input variance irrespective of its semantic content, a known liability in modalities dominated by speckle,\cite{balestriero2024learning} and latent prediction suppresses unpredictable components at the cost of learning its own target, a process with documented sample-complexity requirements.\cite{van2026joint,littwin2024jepa}
Ultrasound foundation models have adopted these objectives in isolation. USCL pretrains a contrastive backbone from ultrasound video and reports strong fine-tuning accuracy on POCUS.\cite{chen2021uscl} USFM and USF-MAE scale masked image modelling to large multi-organ corpora.\cite{jiao2024usfm,megahed2026usf} More recent work transfers latent prediction to ultrasound video and echocardiography.\cite{ellis2026self,mishra2026self,radhachandran2026us} However, as these systems differ simultaneously in corpus, architecture and evaluation, their published numbers do not support inference about the relative merit of the underlying objectives.

Comparative studies have begun to address this. Ivezi\'c et al.\cite{ivezic2026pretext} contrast MAE \cite{he2022masked}, DINOv3 \cite{simeoni2025dinov3} and I-JEPA \cite{assran2023self} on ultrasound and histopathology and conclude that the preferred pretext task is determined by whether a diagnostically relevant signal is spatially localised or globally structured. Complementary evidence indicates that in-domain SSL pretraining can yield large in-domain gains while degrading substantially elsewhere,\cite{anton2022well} and that SSL robustness under distributional change is weaker than natural-image benchmarks suggest.\cite{fedorov2021tasting} Three gaps remain for LUS: existing comparisons operate on static frames rather than temporal video, pretrain on large multi-organ corpora rather than a modest in-domain dataset from a single clinical application, and do not test transfer to an independently acquired LUS dataset.

This work addresses these gaps directly. Our contributions are threefold. 
\ding{182} We conduct a controlled comparison of contrastive, masked, and latent prediction pretraining for LUS video, while holding the backbone, pretraining corpus, optimisation schedule, and frozen evaluation protocol fixed, so that measured differences can be attributed to the pretext objective. 
\ding{183} We pretrain on COVID-BLUeS and evaluate label efficiency at four annotation budgets on POCUS using patient-level five-fold cross-validation and on the independently acquired Mendeley-Uganda dataset.
\ding{184} At the full label budget under linear probing, the ranking on POCUS reverses on Mendeley-Uganda: contrastive pretraining is weakest on POCUS but strongest on Mendeley-Uganda, while latent prediction performs well on POCUS and drops to near-chance performance on Mendeley-Uganda. These results show that POCUS probe accuracy alone does not predict cross-dataset transfer.

\section{METHOD}
\label{sec:method}

\subsection{Pretext objectives}
\begin{figure}[t]
  \centering
  \includegraphics[width=\textwidth]{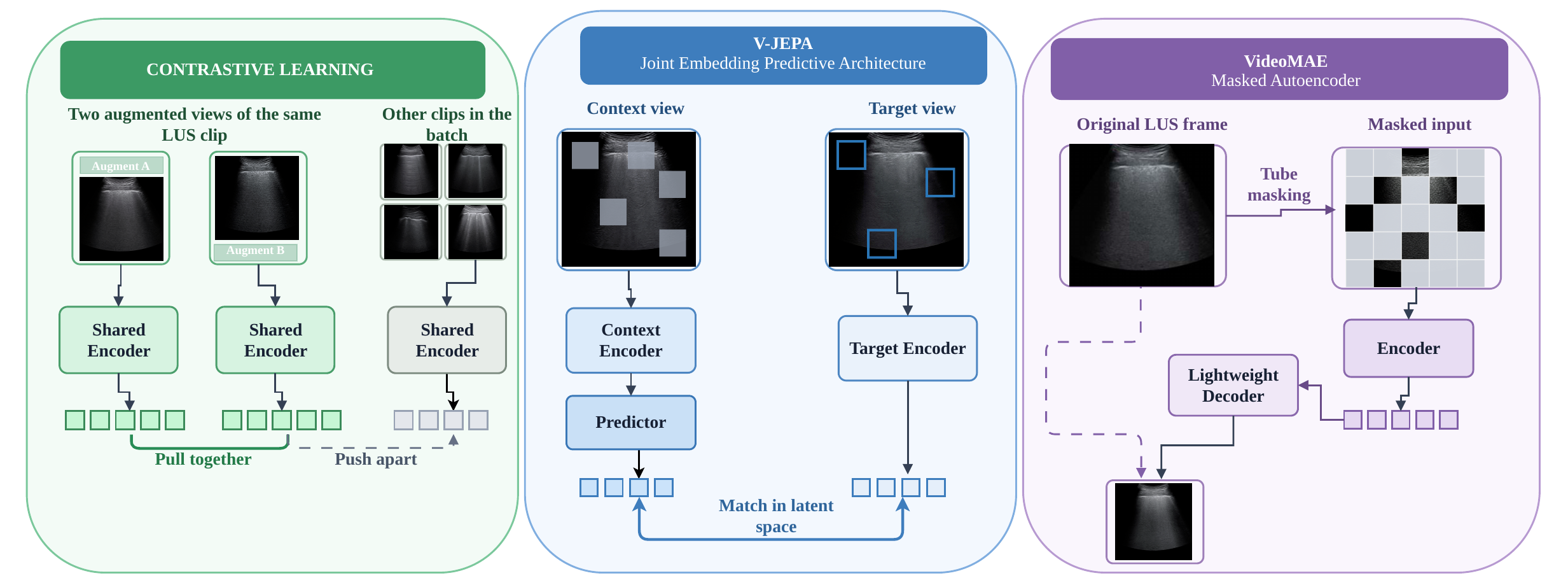}
  \caption{The three pretext objectives applied to the same lung ultrasound clip. The objectives differ in the space in which the loss is evaluated: contrastive learning compares whole-clip embeddings on a hypersphere, masked reconstruction compares images, and latent prediction compares feature tokens.}
  \label{fig:paradigms}
\end{figure}

Let $x$ denote an LUS clip and $f_\theta$ the encoder under study. The paradigms differ only in the pretext task imposed on $f_\theta$, and in particular in the space in which the loss is evaluated (Fig.~\ref{fig:paradigms}).

\textbf{Contrastive.} Two temporally offset, augmented clips $v_1, v_2$ sampled from $x$ are encoded by $f_\theta$ and a momentum encoder $f_\xi$, projected onto a hypersphere, and optimised with InfoNCE so that clips from the same video are aligned and clips from distinct videos are separated. We instantiate this family with the MoCo~v3 objective\cite{chen2021empirical} adapted to video following the spatiotemporal contrastive recipe of VideoMoCo.\cite{pan2021videomoco}

\textbf{Masked reconstruction.} Tube masking removes a fraction $\rho = 90\%$ of spatio-temporal tubelets. The encoder processes only visible tubelets and a lightweight decoder, discarded after pretraining, regresses the withheld pixels under a mean squared error criterion. We instantiate this family with VideoMAE.\cite{tong2022videomae}

\textbf{Latent prediction.} The clip is partitioned into a visible context and masked target blocks. A predictor $g_\varphi$ maps the context representation to an estimate of the target representation, where targets are produced by an exponential moving average encoder $f_\xi$ under a stop-gradient, and the loss is an $L_1$ distance in feature space. We instantiate this family with V-JEPA.\cite{bardes2024vjepa,assran2025v}

\subsection{Controlled pretraining}
All paradigms employ a ViT-S/16 backbone with tubelet size $2$ over $16$-frame clips, are initialised randomly without ImageNet weights, and share optimiser, learning-rate schedule and epoch budget. Pretraining uses the LUS videos from the COVID-BLUeS dataset.\cite{wiedemann2025covid} We use this modest-scale dataset to compare the three objectives under a pretraining budget representative of a single clinical application.

\subsection{Frozen evaluation}
Pretrained encoders are frozen and evaluated with a linear classifier, a $k$NN classifier and an attentive probe, each trained at label budgets of 5\%, 10\%, 50\% and 100\%. The downstream task is three-class classification on POCUS\cite{born2021accelerating} (COVID-19, bacterial pneumonia and healthy) under patient-level five-fold cross-validation. For each fold, the probe is trained on four POCUS folds and evaluated on both the held-out POCUS fold and the full Mendeley-Uganda dataset\cite{katumba2025dataset}. Mendeley-Uganda is excluded from both SSL pretraining and probe fitting and was acquired under a different imaging distribution. We refer to POCUS as in-distribution with respect to probe training, while Mendeley-Uganda serves as the external evaluation dataset. Balanced accuracy is reported as the mean $\pm$ standard deviation over the five POCUS folds and the five corresponding Mendeley-Uganda evaluations. Patient-level partitioning is essential because temporally adjacent LUS frames are near-duplicates, and frame-level partitioning can inflate measured performance.

\section{PRELIMINARY RESULTS}
\label{sec:results}

\begin{table}[t]
\caption{Balanced accuracy (\%) on POCUS and Mendeley-Uganda. Results are reported as mean (standard deviation) over five evaluations. Each POCUS evaluation uses one patient-level test fold. Each Mendeley-Uganda evaluation uses the full external dataset and the probe trained on the corresponding four POCUS training folds. The best value in each column is shown in bold.}
\label{tab:main}
\centering
\scriptsize
\setlength{\tabcolsep}{3.2pt}
\begin{tabular}{llcccccccc}
\toprule
& & \multicolumn{4}{c}{POCUS} & \multicolumn{4}{c}{Mendeley-Uganda} \\
\cmidrule(lr){3-6}\cmidrule(lr){7-10}
Pretraining & Probe & 5\% & 10\% & 50\% & 100\% & 5\% & 10\% & 50\% & 100\% \\
\midrule
\multirow{3}{*}{MoCo v3-S}
& linear
& 45.1\,{\tiny 4.5}
& 41.3\,{\tiny 1.8}
& 41.2\,{\tiny 3.2}
& 42.1\,{\tiny 1.2}
& 61.3\,{\tiny 2.6}
& 62.6\,{\tiny 1.8}
& 62.3\,{\tiny 1.1}
& 62.7\,{\tiny 1.0} \\
& $k$NN
& 33.3\,{\tiny 0.0}
& 33.9\,{\tiny 12.4}
& 44.6\,{\tiny 7.2}
& 37.2\,{\tiny 0.0}
& 57.9\,{\tiny 2.8}
& 57.5\,{\tiny 1.8}
& 57.0\,{\tiny 2.5}
& 57.6\,{\tiny 0.0} \\
& attentive
& \textbf{45.6}\,{\tiny 6.8}
& 39.2\,{\tiny 3.5}
& 42.1\,{\tiny 1.3}
& 41.1\,{\tiny 0.9}
& \textbf{63.0}\,{\tiny 1.6}
& \textbf{64.2}\,{\tiny 0.9}
& \textbf{64.4}\,{\tiny 0.4}
& \textbf{64.4}\,{\tiny 0.2} \\
\midrule
\multirow{3}{*}{VideoMAE-S}
& linear
& 36.5\,{\tiny 17.2}
& 39.7\,{\tiny 7.6}
& 62.9\,{\tiny 12.5}
& \textbf{66.5}\,{\tiny 13.1}
& 42.5\,{\tiny 6.3}
& 45.9\,{\tiny 9.1}
& 51.4\,{\tiny 4.1}
& 53.8\,{\tiny 2.8} \\
& $k$NN
& 38.6\,{\tiny 12.7}
& 42.1\,{\tiny 7.7}
& 51.3\,{\tiny 4.9}
& 60.1\,{\tiny 6.4}
& 38.6\,{\tiny 7.8}
& 38.7\,{\tiny 5.4}
& 39.9\,{\tiny 8.0}
& 44.2\,{\tiny 1.6} \\
& attentive
& 34.7\,{\tiny 15.7}
& \textbf{47.8}\,{\tiny 8.8}
& 62.1\,{\tiny 17.6}
& 60.9\,{\tiny 10.6}
& 40.8\,{\tiny 11.2}
& 45.8\,{\tiny 9.0}
& 51.4\,{\tiny 4.6}
& 48.2\,{\tiny 3.6} \\
\midrule
\multirow{3}{*}{V-JEPA-S}
& linear
& 42.5\,{\tiny 13.8}
& 37.5\,{\tiny 23.6}
& \textbf{64.4}\,{\tiny 11.1}
& 65.4\,{\tiny 11.7}
& 35.2\,{\tiny 9.1}
& 30.3\,{\tiny 10.8}
& 35.4\,{\tiny 7.0}
& 35.1\,{\tiny 4.9} \\
& $k$NN
& 39.9\,{\tiny 12.6}
& 42.5\,{\tiny 15.0}
& 53.6\,{\tiny 10.9}
& 54.0\,{\tiny 13.5}
& 33.7\,{\tiny 6.8}
& 34.1\,{\tiny 7.5}
& 37.7\,{\tiny 6.4}
& 37.7\,{\tiny 2.3} \\
& attentive
& 42.1\,{\tiny 16.4}
& 41.3\,{\tiny 14.5}
& 49.4\,{\tiny 10.1}
& 61.8\,{\tiny 11.5}
& 34.5\,{\tiny 9.0}
& 40.1\,{\tiny 10.2}
& 40.9\,{\tiny 6.4}
& 37.3\,{\tiny 4.3} \\
\bottomrule
\end{tabular}
\end{table}

\begin{figure}[t]
  \centering
  \includegraphics[width=0.92\textwidth]{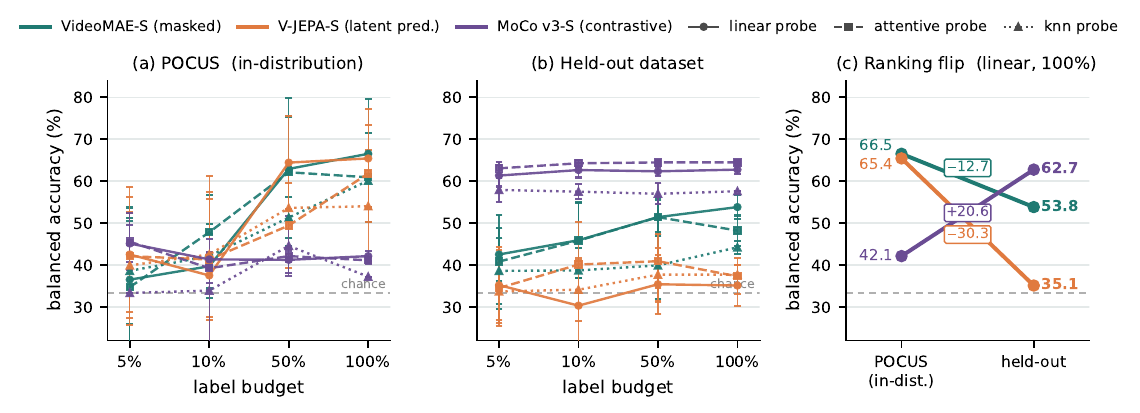}
  \caption{Label efficiency and cross-dataset performance. (a, b) Balanced accuracy across four label budgets using linear, attentive, and $k$NN probes on POCUS and Mendeley-Uganda. On POCUS, VideoMAE and V-JEPA outperform MoCo at the 50\% and 100\% label budgets, although their relative ranking depends on the probe. On Mendeley-Uganda, MoCo performs best for every probe and label budget. (c) At the full label budget under linear probing, MoCo increases by 20.6 points from POCUS to Mendeley-Uganda, while VideoMAE and V-JEPA decrease by 12.7 and 30.3 points, respectively. Error bars in (a, b) show the standard deviation across five POCUS test folds and the five corresponding Mendeley-Uganda evaluations.}
  \label{fig:results}
\end{figure}
Table~\ref{tab:main} and Fig.~\ref{fig:results} report balanced accuracy for the three pretraining objectives. At the full label budget under linear probing, VideoMAE and V-JEPA achieve $66.5 \pm 13.1$ and $65.4 \pm 11.7$ on POCUS, while MoCo achieves $42.1 \pm 1.2$. On Mendeley-Uganda, MoCo performs best at $62.7 \pm 1.0$, followed by VideoMAE at $53.8 \pm 2.8$ and V-JEPA at $35.1 \pm 4.9$. From POCUS to Mendeley-Uganda, MoCo increases by $20.6$ points, while VideoMAE and V-JEPA decrease by $12.7$ and $30.3$ points, respectively (Fig.~\ref{fig:results}c). The ranking therefore changes from VideoMAE $\approx$ V-JEPA $>$ MoCo on POCUS to MoCo $>$ VideoMAE $>$ V-JEPA on Mendeley-Uganda. Under linear probing, MoCo also shows the lowest variability on Mendeley-Uganda at the higher label budgets, with standard deviations of $1.1$ and $1.0$ points at the 50\% and 100\% label budgets, respectively.

\textbf{Interpretation.} The results suggest a trade-off between performance on the POCUS task and transfer to a different acquisition setting. V-JEPA performs well on POCUS but drops to near-chance performance on Mendeley-Uganda. This may be related to the sample-complexity requirements of JEPA-style objectives. Recent ultrasound work has also used static teachers to stabilise JEPA training.\cite{radhachandran2026us} MoCo performs poorly on POCUS but transfers best to Mendeley-Uganda, possibly because its augmentation-based objective encourages invariance across acquisition conditions. VideoMAE falls between the two methods. These results may depend on the size of the pretraining dataset and the distribution used for evaluation.

\section{SCOPE OF THE PRESENT STUDY AND PLANNED ANALYSIS}
\label{sec:planned}

These results are preliminary in two respects: we evaluate only one backbone scale (ViT-S), and the downstream task is limited to three-class classification, which does not directly test the temporal information that motivates video pretraining. The pretraining and evaluation datasets are separate: COVID-BLUeS is used for pretraining, POCUS is used for patient-level cross-validation, and Mendeley-Uganda is used for external evaluation. POCUS is treated as in-distribution with respect to probe training, not SSL pretraining.

The ranking changes across the two evaluation datasets. The full paper will examine this difference using analyses beyond downstream accuracy. Our central planned analysis is a feature-space visualisation of the three pretrained encoders, following Ivezi\'c et al.:\cite{ivezic2026pretext} per-head attention maps, principal-component projections of the patch tokens, and cosine-similarity maps with respect to anchor patches placed on the pleural line and within B-line artefacts, which together reveal whether an objective allocates capacity to diagnostic structure or to speckle and acquisition artefacts. Because POCUS provides frame-level annotations of A-lines, B-lines, consolidation and pleural effusion,\cite{born2021accelerating} we will further score the agreement between these similarity maps and the annotated regions, turning the qualitative maps into a quantitative measure.

\section{CONCLUSION}
\label{sec:conclusion}
We compared three self-supervised pretraining objectives for lung ultrasound using the same encoder backbone, pretraining dataset, training schedule and frozen evaluation protocol. At the full label budget under linear probing, VideoMAE and V-JEPA perform best on POCUS, while MoCo performs best on Mendeley-Uganda and V-JEPA drops to near-chance performance. This reversal shows that POCUS probe accuracy alone is not sufficient for selecting the objective that transfers best to a new acquisition setting. Ongoing work adds the representation-level analyses described in Sec.~\ref{sec:planned} to examine the differences between the learned representations.

\acknowledgments 
This work was supported by the Canadian Foundation for Innovation-John R. Evans Leaders Fund (CFI-JELF) program grant number 42816. Mitacs Accelerate program grant number AWD024298-IT33280. We also acknowledge the support of the Natural Sciences and Engineering Research Council of Canada (NSERC), [RGPIN-2023-03575]. Cette recherche a été financée par le Conseil de recherches en sciences naturelles et en génie du Canada (CRSNG), [RGPIN-2023-03575].    

\bibliography{ref} 
\bibliographystyle{spiebib} 

\end{document}